\documentclass[10pt,twocolumn,letterpaper]{article}

\usepackage{cvpr}
\usepackage{times}
\usepackage{epsfig}
\usepackage{graphicx}
\usepackage{amsmath}
\usepackage{amssymb}
\usepackage{fancyhdr}

\usepackage{booktabs}
\usepackage{algorithm2e}
\usepackage{tensor}
\usepackage{subfigure}

\usepackage[normalem]{ulem} 

\usepackage[pagebackref=true,breaklinks=true,letterpaper=true,colorlinks,bookmarks=false]{hyperref}

\usepackage{subfigure}
\usepackage{float}
\usepackage{nth}
\usepackage{color}
\usepackage{amsmath}
\usepackage{amssymb}
\usepackage{bm}
\usepackage{graphicx}
\usepackage{empheq} 
\usepackage{xcolor}

\newenvironment{packed_itemize}
{\begin{itemize}
    \setlength{\itemsep}{1pt}
    \setlength{\parskip}{0pt}
    \setlength{\parsep}{0pt}
}{\end{itemize}}
\newcommand\tabhead[1]{\textbf{#1}}
\newcommand\parahead[1]{\textbf{#1}:\ }

\newcommand{\C}{\mathcal{C}}

\newcommand{\M}{\mathbf{M}}
\newcommand{\MU}{\boldsymbol{\mu}}
\newcommand{\MUw}{\boldsymbol{\mu}_h}

\newcommand{\SI}{\boldsymbol{\Sigma}}

\newcommand{\Sw}{\boldsymbol{\Sigma}_h}

\newcommand{\e}{\mathbf{e}}
\newcommand{\St}{\mathbf{S}} 
\newcommand{\vect}[1]{\mathbf{#1}}

\newcommand{\ctignore}[1]{{}}

\cvprfinalcopy 

\def\cvprPaperID{007} 

\ifcvprfinal\fi

\usepackage{fancyhdr}
\begin{document}

\title{Real-time Hand Tracking Using a Sum of Anisotropic Gaussians Model}

\author{Srinath Sridhar\textsuperscript{1}, Helge Rhodin\textsuperscript{1},
  Hans-Peter Seidel\textsuperscript{1}, Antti Oulasvirta\textsuperscript{2},
  Christian Theobalt\textsuperscript{1}
\and
\ \\
\textsuperscript{1}Max Planck Institute for Informatics\\
Saarbr\"ucken, Germany\\
{\tt\small \{ssridhar,hrhodin,hpseidel,theobalt\}@mpi-inf.mpg.de}
\and
\ \\
\textsuperscript{2}Aalto University\\
Helsinki, Finland\\
{\tt\small antti.oulasvirta@aalto.fi}
}

\maketitle
\thispagestyle{fancy}

\begin{abstract}
Real-time marker-less hand tracking is of increasing importance in human-computer interaction. 
Robust and accurate tracking of arbitrary hand motion is a challenging problem due to the many degrees of freedom, frequent self-occlusions, fast motions, and uniform skin color.
In this paper, we propose a new approach that tracks the full skeleton motion of the hand from multiple RGB cameras in real-time.
The main contributions include a new generative tracking method which employs an implicit hand shape representation based
on \emph{Sum of Anisotropic Gaussians} (SAG), and a pose fitting energy that is smooth and analytically differentiable making fast gradient based pose optimization possible.
This shape representation, together with a full perspective projection model, enables more accurate hand modeling than a related baseline method from literature.
Our method achieves better accuracy than previous methods and runs at $25$~fps.
We show these improvements both qualitatively and quantitatively on publicly available datasets.
\end{abstract}

\section{Introduction}
Marker-less articulated hand motion tracking has important applications in human--computer interaction (HCI).
Tracking all the degrees of freedom (DOF) of the hand for such applications is hard because of frequent self-occlusions, fast motions, limited field-of-view, uniform skin color, and noisy data.
In addition, these applications impose constraints on tracking, including the need for \emph{real-time} performance, high \emph{accuracy}, \emph{robustness}, and low \emph{latency}.
Most approaches from the literature thus frequently fail on even moderately fast and complex hand motion.  
\begin{figure}
\centering
   \subfigure{\includegraphics[width=0.23\textwidth]{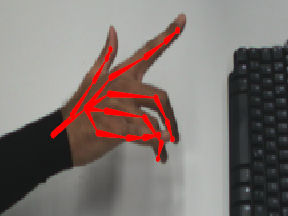}}
   \subfigure{\includegraphics[width=0.23\textwidth]{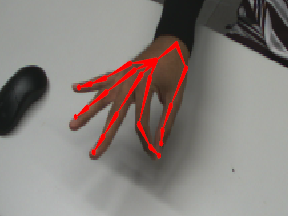}}
   \caption{Qualitative tracking results from our SAG-based tracking method. We achieve a framerate of $25$ fps which is suitable for interaction applications.}
 \label{fig:teaser}
\end{figure}

Previous methods for hand tracking can be broadly classified into either \emph{generative} methods~\cite{oikonomidis_efficient_2011,stenger_model-based_2006,maccormick_partitioned_2000}
or \emph{discriminative} methods~\cite{athitsos_estimating_2003,wang_real-time_2009,wang_6d_2011,keskin_hand_2012}.
Generative methods usually employ a dedicated model of hand shape and articulation whose pose parameters are optimized to fit image data. 
While this yields temporally smooth solutions, real-time performance necessitates fast local optimization strategies which may converge to erroneous local pose optima. 
In contrast, discriminative methods detect hand pose from image features, \eg, by retrieving a plausible hand configuration from a learned space of poses, but the results are usually temporally less stable.

Recently, a promising hybrid method has been proposed that combines generative and discriminative pose estimation for hand tracking from multi-view 
video and a single depth camera~\cite{sridhar_interactive_2013}.
In their work, generative tracking is based on an implicit Sum of Gaussians (SoG) representation of the hand, and discriminative tracking uses a linear SVM classifier to detect fingertip locations.
This approach showed increased tracking robustness compared to prior work but was limited to using isotropic Gaussian primitives to model the hand.

In this paper, we build on this previous method and further develop it to enable fast, more accurate, and robust articulated
hand tracking at real-time rates of $25$ fps.
We contribute a fundamentally extended generative tracking algorithm based on an augmented implicit shape representation. 

The original SoG model is based on the simplifying assumption that all Gaussians in 3D have \textbf{isotropic covariance},
facilitating simpler projection and energy computation. However, in the case of hand tracking this isotropic 3D SoG model reveals several disadvantages. 
Therefore we introduce a new 3D \emph{Sum of Anisotropc Gaussians} (SAG) representation (Figure~\ref{fig:handmodel}) that uses \textbf{anisotropic} 3D Gaussian primitives attached to a kinematic 
skeleton to approximate the volumetric extent and motion of the hand. This step towards a more general class of 3D functions complicates the projection from 3D to 2D and thus the 
computation of the pose fitting energy. 
However, it maintains important smoothness properties and enables a better approximation of the hand shape with less primitives (visualized as ellipsoids in Figure~\ref{fig:handmodel}).
Our approach, in contrast to previous methods~\cite{stoll_fast_2011,sridhar_interactive_2013}, models the full perspective projection of 3D Gaussians.
To summarize, the primary contributions of our paper are:
\begin{packed_itemize}
   \item An advancement of~\cite{sridhar_interactive_2013} that generalizes
     the SoG-based tracking to one based on a new 3D Sum of Anisotropic Gaussians (SAG) model, thus enabling tracking using fewer primitives.
   \item Utilization of a full perspective projection model for projection of 3D Gaussians to 2D in matrix-vector form.
   \item Analytic derivation of the gradient of our pose fitting energy, which is smooth and differentiable, to enable real-time optimization.
\end{packed_itemize}

We evaluate the improvements enabled by SAG-based generative tracking over previous work.
Our contributions not only lead to more accurate and robust real-time tracking but also allow tracking of objects in addition to the hand.
\section{Previous Work}\label{sec:litrev}
Following the survey of Erol \etal~\cite{erol_vision-based_2007} we review previous work by categorizing them into either \emph{model-based tracking} methods or \emph{single frame pose estimation} methods.
Model-based tracking methods use a hand model, usually a kinematic skeleton with additional surface modeling, to estimate the parameters
that best explain temporal image observations. Single frame methods are more diverse in their algorithmic recipes, they make fewer assumptions
about temporal coherence and often use non-parametric models of the hand.
Hand poses are inferred by exploiting some form of inverse mapping from image features to a space of hand configurations.

\parahead{Model-based Tracking}
Rehg and Kanade~\cite{rehg_visual_1994} were one of the first to present a kinematic model-based hand tracking method.
Lin \etal~\cite{lin_modeling_2000,wu_capturing_2001} studied the
constraints of hand motion and proposed feasible base states to reduce the search space size.
Oikonomidis \etal~\cite{oikonomidis_efficient_2011} presented a method based on particle swarm optimization
for full DoF hand tracking using a depth sensor and achieved a frame rate of $15$
fps with GPU acceleration.
Other model-based methods using global optimization for pose inference fail to perform at real-time frame rates~\cite{stenger_model-based_2006,maccormick_partitioned_2000}.

Primitive shapes such as spheres and (super-)quadrics have been explored for tracking objects~\cite{krivic_contour_2003}, 
and, recently, for tracking hands~\cite{qian_realtime_2014}.
However, perspective projection of complex shapes is hard to represent analytically and therefore fast optimization is hard.
In this work we use anisotropic Gaussian primitives with analytical expression for perspective projection.
An overview of perspective projection of spheroids, which are conceptually similar to anisotropic Gaussians, can be found in ~\cite{eberly_perspective_1999}.

Tracking hands with objects imposes additional constraints on hand motion. Methods proposed by Hamer \etal~\cite{hamer_tracking_2009,hamer_object-dependent_2010},
and others~\cite{oikonomidis_full_2011,romero_hands_2010,ballan_motion_2012} model these constraints.
However, these methods require offline computation and are unsuitable for interaction applications.

\parahead{Single Frame Pose Estimation}
Single frame methods estimate hand pose in each frame of the input sequence without taking temporal information into account.
Some methods build an exemplar pose database and formulate pose estimation as a database indexing problem~\cite{athitsos_estimating_2003}.
The retrieval of the whole hand pose was explored by Wang and Popovi\'{c}~\cite{wang_real-time_2009,wang_6d_2011}.
However, the hand pose space is large and it is difficult to sample it with sufficient granularity for jitter-free pose estimation.
Sridhar \etal~\cite{sridhar_interactive_2013} proposed a part-based pose retrieval method to reduce the search space. 
Decision and regression forests have been successfully used in full body pose estimation to learn human pose from a large synthetic dataset~\cite{shotton_real-time_2011}.
This approach has been recently adopted for hands~\cite{keskin_hand_2012,tang_real-time_2013,xu_efficient_2013,tang_latent_2014}.
These methods generally lack temporal stability and recover only joint positions or part labels instead of a full kinematic skeleton.

\parahead{Hybrid Tracking}
Hybrid frameworks that combine the advantages of model-based tracking and single frame
pose estimation can be found in full body pose estimation~\cite{ye_accurate_2011,baak_data-driven_2011,wei_accurate_2012} and early hand tracking~\cite{rosales_combining_2006}.
A hybrid method that uses color and depth data for hand tracking was proposed~\cite{sridhar_interactive_2013}.
However, this method is limited to studio conditions and uses isotropic Gaussian primitives.
In this paper, we extend their method by introducing an improved (model-based) generative tracker.
This new tracker alone leads to higher tracking accuracy and robustness than the baseline method it extends.

\section{Tracking Overview}\label{sec:hybrid}
\begin{figure*}[ht!]
\centering
\includegraphics[width=\textwidth]{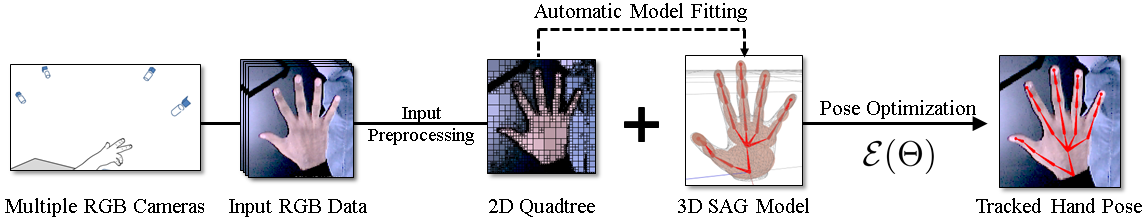}
   \caption{Overview of the our tracking framework. We present a novel generative tracking method that models the hand as a \emph{Sum of Anisotropic Gaussians}.
     We obtain better accuracy and robustness than previous work.}
 \label{fig:hybrid_tracking}
\end{figure*}

Figure~\ref{fig:hybrid_tracking} shows an overview of our tracking framework.
The goal is to estimate hand pose robustly by maximizing the similarity between the hand model and the input images.
The tracker developed in this paper extends the generative pose estimation method of the tracking algorithm in~\cite{sridhar_interactive_2013}.

The input to our method are a set of RGB images from $5$ video cameras (Point Grey Flea 3) of resolution $320 \times 240$ (see Figure~\ref{fig:hybrid_tracking}).
The cameras are calibrated and run at $60$~fps.
The hand is modeled as a full kinematic skeleton with $26$ degrees-of-freedom (DOF),
and unlike other methods that deliver only joint locations or part labels in
the images, our approach computes full kinematic joint
angles $\Theta^* = \{\theta_j^*\}$.

Our method maximizes the similarity between the 3D hand
model projected into all RGB images, and the images themselves, by means of a fast iterative
local optimization algorithm.
The main novelty and contribution of this work is the new shape
representation and pose optimization framework used in the tracker (Section~\ref{sec:generative}).
Our pose fitting energy is smooth, differentiable and allows fast gradient-based optimization.
This enables real-time hand tracking with higher accuracy and robustness while using fewer Gaussian primitives.

\section{SAG-based Generative Tracking}\label{sec:generative}
The generative tracker from~\cite{sridhar_interactive_2013} is based on a representation called 
a Sum of Gaussians (SoG) model, that was originally proposed in~\cite{stoll_fast_2011} for full body tracking. 
The basic concept of the SoG model is to approximate the 3D volumetric extent of the hand by \textbf{isotropic Gaussians} attached to  
the bones of the skeleton, with a color associated to each Gaussian (Figure~\ref{fig:handmodel} (c)). Similarly, input images are segmented to regions of 
coherent color, and each region is approximated by a 2D SoG (Figure~\ref{fig:EllipsoidProjection} (b-c)). A SoG-based pose fitting energy was defined by 
measuring the overlap (in terms of spatial support and color) between the projected 3D Gaussians and all the image Gaussians. This 
energy is maximized with respect to the degrees of freedom to find the correct pose.

Unfortunately, a faithful approximation of the hand volume with a collection of isotropic 3D Gaussians often requires many Gaussian primitives with 
small standard deviation, a problem akin to packing a volume with spherical primitives. With SoG, this leads to sub-optimal hand shape approximation 
and increased computational complexity due to a high number of primitives in 3D (Figure~\ref{fig:handmodel}).
In this paper, we extend the SoG model and represent the hand shape in 3D with \textbf{anisotropic Gaussians}, yielding a \emph{Sum of Anisotropic Gaussians} model (see Figure~\ref{fig:teaser}).
This not only enables a better approximation of the hand shape with less 3D primitives (Figure~\ref{fig:handmodel}), but also leads to higher pose estimation accuracy 
and robustness. The move to anisotropic 3D Gaussians complicates their projection into 2D where scaled orthographic projection~\cite{stoll_fast_2011,sridhar_interactive_2013} cannot be used.
But we show that the numerical benefits of the SoG representation hold equally for the SAG model:
1) We derive a pose fitting energy that is smooth and analytically differentiable for the SAG model under perspective projection
that allows efficient optimization with a gradient-based iterative solver;
2) We show that occlusions can be efficiently approximated with the SAG model within our energy formulation.
This is in contrast to many other generative trackers where occlusion handling leads to discontinuous pose fitting energies.

\subsection{Fundamentals of SAG Model}
We represent both the volume of the hand in 3D, as well as the RGB images with a collection 
of anisotropic Gaussian functions. A \emph{Sum of Anisotropic Gaussians} (SAG) model thus takes the form:
\begin{align}
\mathcal{C}(\mathbf{x}) = \sum\limits_{i=1}^n \mathcal{G}_i(\boldsymbol\mu_i,\boldsymbol\Sigma_i),
\label{eqn:SAG}
\end{align}
where $\mathcal{G}_i(.)$ denotes a un-normalized, anisotropic Gaussian 
\begin{align}
\mathcal{G}(\boldsymbol\mu_i,\boldsymbol\Sigma_i) :=  \exp\left[-\frac{1}{2}({\mathbf x}-{\boldsymbol\mu_i})^T{\boldsymbol\Sigma_i}^{-1}({\mathbf x}-{\boldsymbol\mu_i}) \right],
\end{align}
with mean $\boldsymbol\mu_i$ and covariance matrix is $\boldsymbol\Sigma_i$ for the $i^{th}$ Gaussian. 
Each Gaussian also has an
associated average color vector $\mathbf{c}_i$ in HSV color space.

Using the above representation, we model the hand surface as a sum
of 3D anisotropic Gaussians (3D SAG), where $\mathbf{x} \in \mathbb{R}^3$.
We also approximate the input RGB images as a sum of 2D isotropic Gaussians (2D SoG), where $\mathbf{x} \in \mathbb{R}^2$.
This is an extension of the SoG representation proposed earlier in~\cite{stoll_fast_2011,sridhar_interactive_2013}, which 
was limited to isotropic Gaussians. 

\parahead{3D Hand Modeling}
We model the volumetric extent of the hand as a 3D sum of anisotropic Gaussians model 
(3D SAG), where $\mathbf{x} \in \mathbb{R}^3$. Each $\mathcal{G}_i$ in the 3D SAG is attached to one 
bone of the skeleton, and thus moves with the local frame of the bone (Figure~\ref{fig:handmodel}). 
A linear mapping between skeleton joint angles and a pose parameter space, $\Theta = \{\theta_j\}$, is constructed.
The skeleton pose parameters are further constrained to a predefined range of motion 
reflecting human anatomy,
$\theta_j \in [l_{min}^j, l_{max}^j]$. 
The use of anisotropic Gaussians, whose spatial density is controlled by the covariances,  
enables us to approximate the general shape of the hand with less primitives than needed 
with the original isotropic SoG model (Figure~\ref{fig:handmodel}).
This is because we can create a better \emph{packing} of the hand volume with more generally elongated 
Gaussians, particularly for approximating cylindrical structures like the fingers.
\begin{figure}
\centering
   \subfigure{\includegraphics[height=0.125\textwidth]{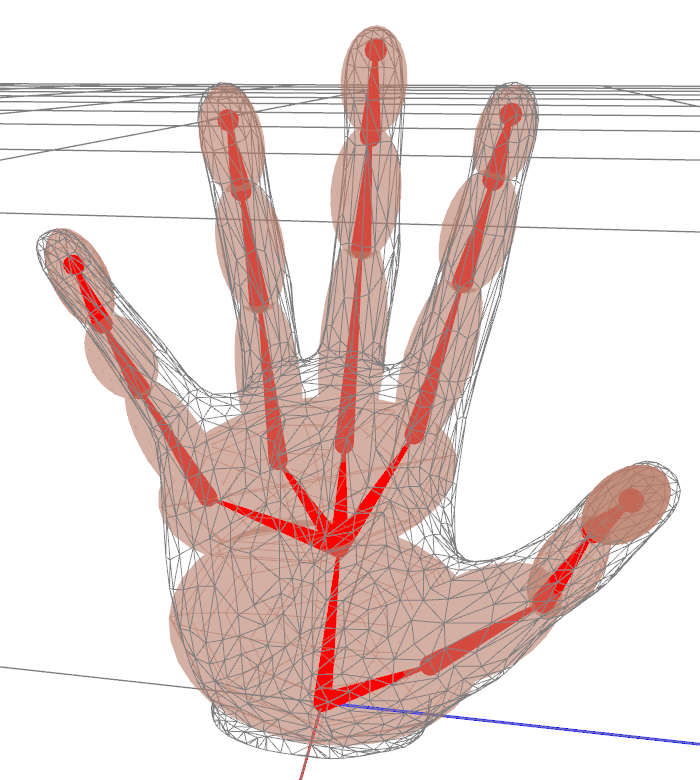}}
   \subfigure{\includegraphics[height=0.125\textwidth]{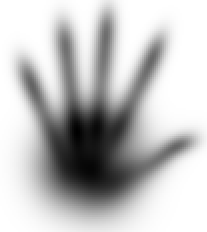}}
   \subfigure{\includegraphics[height=0.125\textwidth]{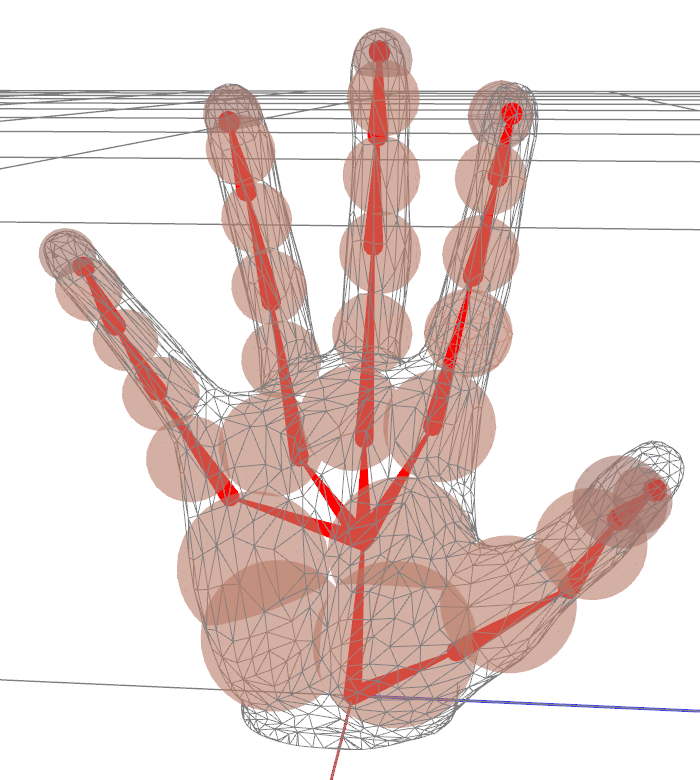}}
   \subfigure{\includegraphics[height=0.125\textwidth]{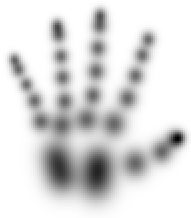}}
   \caption{(a) Our 3D SAG hand model with $17$ \textbf{anisotropic Gaussians} visualized as ellipsoids with radii equal to the standard deviation.
     (c) A 3D SoG hand model with $30$ \textbf{isotropic Gaussians} visualized as spheres.
     (b) Visualizes the SAG model density when projected into 2D; with less primitives,
     the shape of the hand is much better approximated than with the SoG model
     (d)~\cite{sridhar_interactive_2013}.
     \label{fig:handmodel}}
 \label{fig:handmodel}
\end{figure}
The Gaussians in 3D have infinite spatial support, which is an advantageous property for 
pose fitting, as explained later, but also means that the SAG does not represent a finite volume ($\mathcal{C}(\mathbf{x})>0$ everywhere).
We therefore assume that the hand is well modeled by a 3D SAG if the surface passes through each 
Gaussian at a distance of $1$ standard deviation from the mean.
 
\parahead{Hand Model Initialization}
The hand model for tracking requires initialization of the skeleton dimensions, Gaussian covariances 
that control their shapes, and associated colors for an actor before it can be used for tracking.
Our method accepts manually created hand models which could be obtained from a laser scan.
Alternatively, we also provide a fully automatic procedure to obtain a hand model to fit each person.
This method uses a greedy optimization strategy to optimize for a total of $3$ global hand shape parameters
and $3$ independent scaling parameters (along the local principal axes) for each of the $17$ Gaussians in the hand model.
We observed that starting with a manual model and then using the greedy fitting algorithm works best.

\parahead{2D RGB Image Modeling}
We approximate the input RGB images using 2D SoG, $\mathcal{C}_I$,
by quad-tree clustering of regions of similar color.
While it would also be possible to approximate the image as 2D SAG, the 
computational expense of the non-uniform region segmentation would prohibit realtime performance.
We found in our experiments that around $500$ 2D image Gaussians were generated for each camera image.

\subsection{Projection of 3D SAG to 2D SAG}\label{sec:SAG}
Pose optimization (Section~\ref{sec:energy}) requires the comparison of
the projections of the 3D SAG into all camera views, with the 2D SoG of each RGB image. 
Intuitively (for a moment
ignoring infinite support), SAG and SoG can be visualized as ellipsoids
and spheres, respectively. 

The perspective projections of spheres and ellipsoids both yield ellipses
in 2D~\cite{eberly_perspective_1999}. For the case of isotropic Gaussians
in 3D, like in the earlier SoG model, projection of a 3D Gaussian can
be approximated as a 2D Gaussian with a standard deviation that is a scaled
orthographic projection of the 3D standard deviation~\cite{sridhar_interactive_2013,stoll_fast_2011}.
This simple approximation does not hold for our anisotropic Gaussians.   

\begin{figure}
\centering
\includegraphics[width=0.47\textwidth]{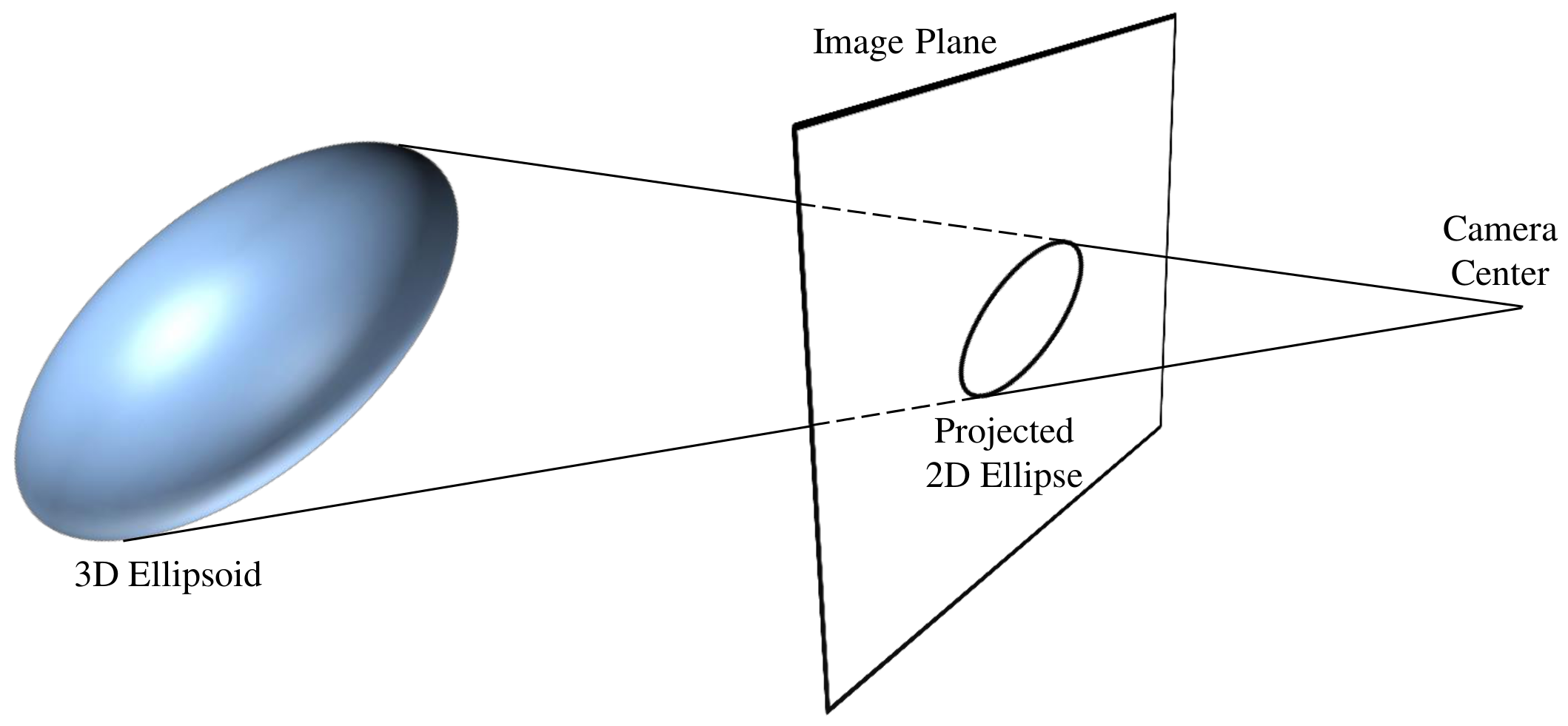}
   \caption{Sketch of the perspective projection of an ellipsoids as the intersection 
 of the image plane with the cone formed by the camera center and the ellipsoid.} 
 \label{fig:EllipsoidProjection}
\end{figure}

Therefore, we utilize an exact perspective projection $\Pi$ of ellipsoids, in order to 
model the projection of the 3D SAG hand model, $\mathcal{C}_H$, into its 2D SAG equivalent, $\mathcal{C}_P$.
Given an ellipsoid in $\mathbb{R}^3$ with associated mean, $\MU_h$ and
covariance matrix, $\SI_h$, its perspective projection can be visualized as an ellipse in $\mathbb{R}^2$
with parameters $\SI_p$ and mean $\MU_p$. Figure \ref{fig:EllipsoidProjection} (a) sketches the perspective projection, intuitively, $\Pi$ can be thought of as the intersection of
the elliptical cone (formed by the ellipsoid and the camera center) with the image plane.
Without loss of generality, we assume the camera to be at the origin
with a camera matrix, $\mathbf{P} = \mathbf{K}\,[\,\mathbf{I}\,|\,\mathbf{0}\,]$.
The parameters of the projected Gaussian are given by
\begin{align}
  \MU_p = & \frac{1}{|\mathbf{M}_{33}|} \, \mathbf{K}_{33} \left[
    \begin{array}{c}
     \phantom{-} |\mathbf{M}_{31}| \\
      -|\mathbf{M}_{23}|
    \end{array}
    \right] + 
    \left[
    \begin{array}{c}
      k_{13} \\
      k_{23}
    \end{array}
    \right]
, \ \\
\mathbf{\Sigma}_p = & -\frac{|\mathbf{M}|}{|\mathbf{M}_{33}|} \, \mathbf{K}_{33} \, \mathbf{M}_{33}^{-1} \, \mathbf{K}_{33}^T,
\end{align}
where
\begin{align}
  \mathbf{M} = \SI_h^{-1} \, \MU_h \MU_h^T \SI_h^{-\top} - \left(\MU_h^\top \SI_h^{-1} \MU_h - 1 \right) \SI_h^{-1},
\end{align}
$|\mathbf{M}|$ is the determinant of $\mathbf{M}$, $\mathbf{A}_{ij}$
is a matrix $\mathbf{A}$ with its $i^{th}$ row and $j^{th}$ column removed,
and $k_{ij}$ is the element at the $i^{th}$ row and $j^{th}$ column of $\mathbf{K}$.
Please see the supplementary material for
the derivation of $\mathbf{M}$ and the projection with arbitrary camera
matrices. This more general projection also leads to a more involved 
pose fitting energy with more involved derivatives than for the SoG model, as explained in the next section. 

\subsection{Pose Fitting Energy}\label{sec:energy}
We now define an energy that measures the quality of overlap between the projected 3D SAG $\mathcal{C}_P = \Pi(\mathcal{C}_H)$, 
and the image SoG $\mathcal{C}_I$, and that is optimized with respect to the pose parameters $\Theta$ of the hand model. 
Our overlap measure is an extension of the SoG overlap measure~\cite{stoll_fast_2011} to a SAG. 
Intuitively, we assume that two Gaussians in 2D match well if their spatial support aligns, and their 
color matches. This criterion can be expressed by the spatial integral over their product, weighted by a color similarity 
term.
The similarity of any two sets, $\C_a$ and $\C_b$, of SAG or SoG in 2D (including combined models with both isotropic and anisotropic Gaussians)
can thus be defined as
\begin{align}
E(\C_a,\C_b)
&= \sum_{p \in \C_a} \sum_{q \in \C_b} d(\vect{c}_p,\vect{c}_q) \int_\Omega \mathcal{G}_p (\vect{x}) \mathcal{G}_q (\vect{x}) \, \mathrm{d}\vect{x} \nonumber\\
&= \sum_{p \in \C_a} \sum_{q \in \C_b} d(\vect{c}_p,\vect{c}_q) \, D_{pq} = \sum_{p \in \C_a} \sum_{q \in \C_b} E_{pq}
\label{eqn:weighted_overlap}
\end{align}
where $E_{pq}$ is the integral overlap measure mentioned earlier, $d(\mathbf{c}_p,\mathbf{c}_q)$
measures color similarity using the Wendland function~\cite{wendland_piecewise_1995}, and
$E_{pq} = d(\mathbf{c}_p,\mathbf{c}_q) \, D_{pq}$. Unlike the SoG model,
for the general case of potentially anisotropic Gaussians, the term $D_{pq}$ evaluates to
\begin{align}
D_{pq}
= \frac{\sqrt{(2 \pi)^2 |\boldsymbol\Sigma_p \, \boldsymbol\Sigma_q|}}{\sqrt{|(\boldsymbol\Sigma_p+\boldsymbol\Sigma_q)|}} 
\e^{-\frac{1}{2} (\boldsymbol\mu_p-\boldsymbol\mu_q)^{T}(\boldsymbol\Sigma_p+\boldsymbol\Sigma_q)^{-1}(\boldsymbol\mu_p-\boldsymbol\mu_q)}.
\end{align}
Using this Gaussian similarity formulation allows us to compute the similarity
between the image SoG $\mathcal{C}_I$ and the projected hand SAG $\mathcal{C}_P$.

We also need to consider occlusions of Gaussians from a camera view.
Computing a function that indicates occlusion analytically independent of pose parameters is generally difficult and may lead to a discontinuous 
similarity function. Thus, we use a heuristic approximation of occlusion~\cite{stoll_fast_2011} that yields a continuous fitting energy defined 
as follows   
\begin{align}
 &E_{sim}\left[\mathcal{C}_I, \mathcal{C}_H\right]
 =  \sum_{q \in \mathcal{C}_I} \min \left( \sum_{p \in{ \Pi(\mathcal{C}_H)}} \, w_p^h \, E_{pq}, E_{qq} \right),
 \label{eqn:model_similarity}
\end{align}
where $w_p^h$ is a weighting factor for each projected 3D Gaussian of the hand model. 
$E_{qq}$ is the overlap of an image Gaussian with itself. With this formulation, an image Gaussian cannot contribute more to the overlap 
similarity than by its own footprint in the image.
To find the hand pose, $E_{sim}$ is optimized with respect to $\Theta$, as described in the 
following section. Note that the infinite support of the Gaussians is advantageous
as it leads to an \emph{attracting force} between the projected model and the image of the hand,
even if they do not overlap in a camera view. 

\subsection{Pose Optimization}
The final energy that we maximize to find the hand pose takes the form
\begin{align}
\mathcal{E}(\Theta) = E_{sim}(\Theta) - w_l \, E_{lim}(\Theta),
\label{eqn:energy}
\end{align}
where $E_{lim}(\Theta)$ penalizes motions outside of parameter limits quadratically, and weight $w_l$ is empirically set to $0.1$.
With the SoG formulation, it was possible to express the energy function (with a scaled orthographic projection) in a closed form analytic expression, 
and to derive the analytic gradient.
We have found that $E_{sim}(\Theta)$ in our SAG-based, even with its full perspective projection model, can still be written in closed form
with analytic gradient.

%
%

We derive the analytical gradient of ${E_{sim}}$ with respect to the degrees of freedom $\Theta$ 
 in three steps. For each Gaussian pair $(h,q)$ and parameter $\theta_j$ we compute
{
\footnotesize
\begin{align}
\left(
\frac{\partial \Sw}{\partial \theta_j},\frac{\partial \MUw}{\partial \theta_j} 
\right) \ 
\xrightarrow{a)}
 \ \left(
\frac{d \boldsymbol M}{d \theta_j}
\right)  \ 
\xrightarrow{b)}
 \ \left(
\frac{d \SI_p}{d \theta_j},\frac{d \MU_p}{d \theta_j} 
\right)  \ 
\xrightarrow{c)}
 \  \left(
\frac{d D_{pq}}{d \theta_j}
\right).
\label{eqn:DerivativeDiagram}
\end{align}
}%
We exemplify the computation at hand of step a); the input to a) is the change of the ellipsoid covariance matrix $\partial \Sw^{-1}$ and the change of position $\partial \MUw$ with respect to the DOF $\theta_j$. In this step we are interested in the total derivative
\begin{align}
\frac{d \boldsymbol M}{d \theta_j}  = &\sum_{i \in \{1,2,3\}} \frac{\partial \M}{\partial {\MUw}_i} \frac{\partial {\MUw}_i}{\partial \theta_j}\nonumber \\
& + \sum_{k \in \{1,\cdots,6\}} \frac{\partial \M}{\partial {\Sw}_{k}^{-1}}  \frac{\partial {\Sw}_{k}^{-1}}{\partial \theta_j}.
\end{align}
Following matrix calculus, the partial derivatives of the cone matrix $\boldsymbol M$ with respect to $\MUw,\Sw$ are
{
\begin{align}
\frac{\partial \M}{\partial {\MUw}_i}  
= &\,
\Sw^{-1} (\e^i \MUw^\top + \e^i \MUw^\top)^\top \Sw^{-1}\nonumber \\
& - \left(({\e^i}^\top \Sw^{-1} \MUw) + ({\e^i}^\top \Sw^{-1} \MUw)^\top \right) \Sw^{-1}, \nonumber \\
\frac{\partial \M}{\partial {\Sw}^{-1}_{k} }
= &\, 
H + H^\top - {\MUw}^\top \St^k{\MUw} \Sw^{-1} \nonumber \\
& -( \MUw^\top \Sw^{-1} \MUw - 1) \St^k,
\label{eq:ellipticalConeDerivativeMu}
\end{align}
}%
with $e^i$ the $i^{th}$ unit vector, ${\MUw}_i$ the $i^{th}$ entry of $\MUw$, $\boldsymbol H = \St^k \MUw \MUw^\top \Sw^{-\top}$, where $k$ indexes the unique elements of the symmetric matrix $\Sw^{-1}$, and $\St^k$ is the symmetric structure matrix with the $k^{th}$ elements equal to one, and zero otherwise.
Steps b) and c) are derived in a similar manner and can be found in the supplementary document. The total similarity energy ${E_{sim}}$ is the weighted sum over all $\theta_j$ and $D_{pq}$ according to equation~\ref{eqn:model_similarity}.
Combined with the analytical gradient of $E_{lim}(\Theta)$ we obtain
an analytic formulation for $\frac{\partial}{\partial \Theta} \mathcal{E}(\Theta)$. As sums of independent terms, both $\mathcal{E}$ and $\frac{\partial}{\partial \Theta} \mathcal{E}$ lend themselves to parallel implementation. 

Even though evaluation of fitting energy and gradient is much more involved than for the SoG model, both share the same smoothness properties and can be evaluated efficiently, and thus 
an optimal pose estimate can be computed effectively using a standard gradient-based optimizer.
The optimizer is initiliazed with an extrapolation of the pose parameters from the two previous time steps.
The SAG framework leads to much better accuracy and robustness and requires far fewer shape primitives to be compared, as validated in 
Section~\ref{sec:evaluation}. 


\section{Experiments}\label{sec:evaluation}
We conducted extensive experiments to show that our SAG-based tracker outperforms the SoG based method it extends.
We also compare with another state-of-the-art method that uses a single depth camera~\cite{tang_latent_2014}.
We ran our experiments on the publicly available \textbf{Dexter~1}
dataset~\cite{sridhar_interactive_2013} which has ground truth annotations.
This dataset contains challenging, slow and fast motions.
We processed all $7$ sequences in the dataset and, while \cite{sridhar_interactive_2013} evaluated their
algorithm only on the slow motions, we evaluated our method on fast motions as well.

For all results we used $10$ gradient ascent iterations.
Our method runs at a framerate of $25$ fps on an Intel Xeon E5-1620
running at 3.60~GHz with 16~GB RAM.
Our implementation of the SoG-based tracker of \cite{sridhar_interactive_2013}
runs slightly faster at $40$ fps.

\begin{figure}[!t]
  \centering
  \includegraphics[width=0.4\textwidth]{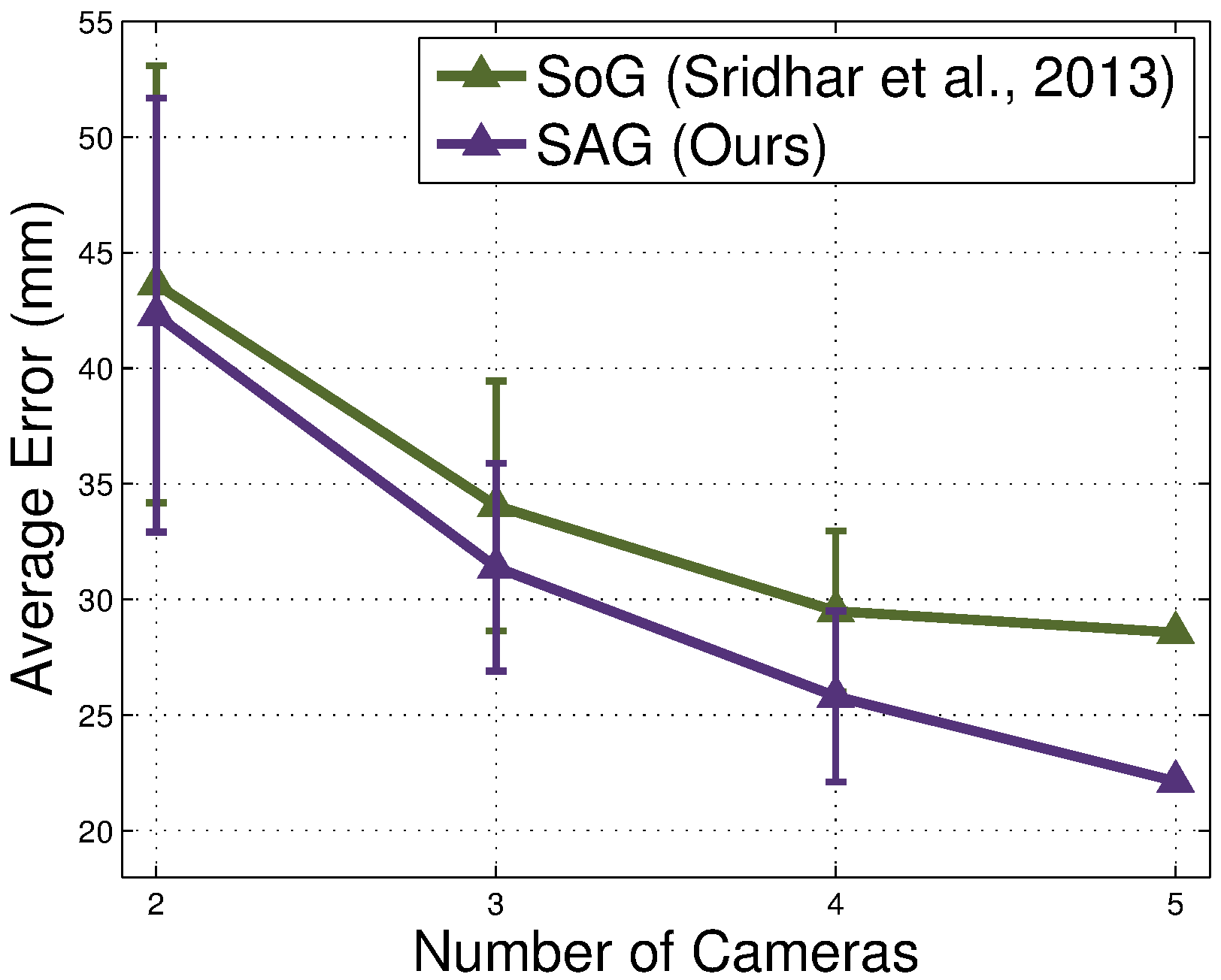}
   \caption{This figure shows a comparison of tracking error for \texttt{SAG} and \texttt{SoG} with $2$ to $5$ cameras.
     A total of $156$ runs were required for \texttt{SAG} and \texttt{SoG} with different camera combinations. The results show that \texttt{SAG} outperforms \texttt{SoG}. Best viewed in color.}
   \label{fig:results2}
\end{figure}

\begin{figure*}[!t]
  \centering
   \includegraphics[width=\textwidth]{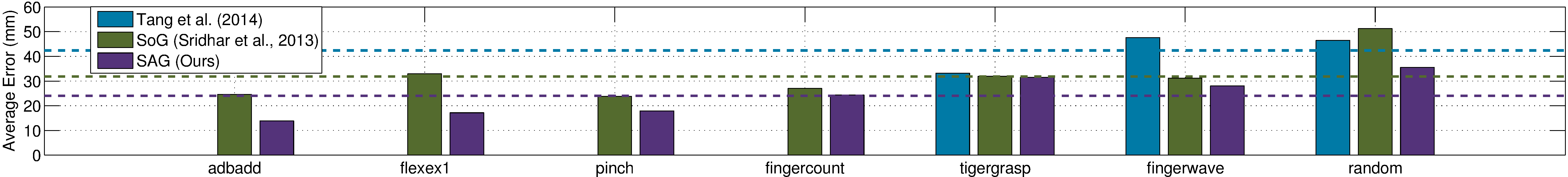}
   \caption{Average errors for all sequences in the Dexter 1 dataset. Our method has the lowest average error of $24.1$~mm compared to SoG ($31.8$~mm) and ~\cite{tang_latent_2014} ($42.4$~mm).
     The dashed lines represent average errors over all sequences. Best viewed in color.}
   \label{fig:results1}
\end{figure*}

\parahead{Accuracy}
Figure~\ref{fig:results1} shows a plot of the average error for each sequence in our dataset.
Over all sequences, \texttt{SAG} had an error of $24.1$~mm, \texttt{SoG} had an error of $31.8$~mm, and \cite{tang_latent_2014} had an error of $42.4$~mm (only 3 sequences).
The mean standard deviations were $11.2$~mm for \texttt{SAG}, $13.9$~mm for \texttt{SoG}, and $8.9$~mm for \cite{tang_latent_2014} (3 sequences only).
Our errors are higher than those reported by
\cite{sridhar_interactive_2013} because we performed our experiments on both
the slow and fast motions as opposed to slow motions only.
Additionally, we discarded the palm center used by \cite{sridhar_interactive_2013}
since this is not clearly defined.
We would like to note that \cite{tang_latent_2014} perform their tracking on the depth data in Dexter 1
, and use no temporal information.
In summary, \texttt{SAG} achieves the lowest error and is $7.7$~mm better than \texttt{SoG}.
This improvement is nearly the width of a finger thus making it a significant gain in accuracy.

\parahead{Error Frequency}
Table~\ref{tab:error_rate} shows an alternative view of the accuracy and robustness improvement of \texttt{SAG}.
We calculated the percentage of frames of each sequence in which the tracking error is less than $x$~mm where $x \in \{15, 20, 25, 30, 45\}$.
This experiment shows clearly that \texttt{SAG} outperforms \texttt{SoG} in almost all sequences and error bounds.
In particular the improvement in {\bf accuracy} is measured by the increased number of frames with error smaller than $15$~mm,
and the {\bf robustness to fast motions} by the smaller number of dramatic failures of errors larger than $30$~mm.
For example, in the \texttt{adbadd} sequence $70.7\%$ of frames are better than $15$~mm for \texttt{SAG} while only $34.5\%$ of frames for \texttt{SoG}.
Note that when $x = 100$~mm, the percentage of frames $< x$~mm is $100\%$ for \texttt{SAG}.
\begin{table*}[t]
\centering
{
\small
  \begin{tabular}{|p{0.18in}||p{0.27in}p{0.27in}||p{0.27in}p{0.27in}||p{0.27in}p{0.27in}||p{0.27in}p{0.27in}||p{0.27in}p{0.27in}||p{0.27in}p{0.27in}||p{0.27in}p{0.27in}|}
    \hline
    \multicolumn{1}{|c||}{\tabhead{Error $<$ $x$~mm}} & \multicolumn{2}{c||}{\tabhead{adbadd}} & \multicolumn{2}{c||}{\tabhead{fingercount}}
    & \multicolumn{2}{c||}{\tabhead{fingerwave}} & \multicolumn{2}{c||}{\tabhead{flexex1}} & \multicolumn{2}{c||}{\tabhead{pinch}}
    & \multicolumn{2}{c||}{\tabhead{random}} & \multicolumn{2}{c|}{\tabhead{tigergrasp}}\\
    \cline{2-15}
     & \multicolumn{1}{c}{SoG} & \multicolumn{1}{c||}{SAG}
     & \multicolumn{1}{c}{SoG} & \multicolumn{1}{c||}{SAG}
     & \multicolumn{1}{c}{SoG} & \multicolumn{1}{c||}{SAG}
     & \multicolumn{1}{c}{SoG} & \multicolumn{1}{c||}{SAG}
     & \multicolumn{1}{c}{SoG} & \multicolumn{1}{c||}{SAG}
     & \multicolumn{1}{c}{SoG} & \multicolumn{1}{c||}{SAG}
     & \multicolumn{1}{c}{SoG} & \multicolumn{1}{c|}{SAG}\\
    \hline
    \multicolumn{1}{|c||}{15} & \multicolumn{1}{r}{34.5} & \multicolumn{1}{r||}{\textbf{70.7}}  & \multicolumn{1}{r}{\textbf{13.1}} & \multicolumn{1}{r||}{8.7}            & \multicolumn{1}{r}{11.0} & \multicolumn{1}{r||}{\textbf{16.7}}  & \multicolumn{1}{r}{ 5.2} & \multicolumn{1}{r||}{\textbf{50.0}}  & \multicolumn{1}{r}{10.8}           & \multicolumn{1}{r||}{\textbf{34.0}}            & \multicolumn{1}{r}{ 3.3} & \multicolumn{1}{r||}{\textbf{10.5}}  & \multicolumn{1}{r}{\textbf{11.2}} & \multicolumn{1}{r|}{10.2}  \\
    \multicolumn{1}{|c||}{20} & \multicolumn{1}{r}{48.1} & \multicolumn{1}{r||}{\textbf{97.5}}  & \multicolumn{1}{r}{\textbf{35.2}} & \multicolumn{1}{r||}{33.4}           & \multicolumn{1}{r}{31.0} & \multicolumn{1}{r||}{\textbf{34.3}}  & \multicolumn{1}{r}{12.1} & \multicolumn{1}{r||}{\textbf{79.4}}  & \multicolumn{1}{r}{30.78}          & \multicolumn{1}{r||}{\textbf{66.3}}            & \multicolumn{1}{r}{ 5.3} & \multicolumn{1}{r||}{\textbf{21.4}}  & \multicolumn{1}{r}{\textbf{25.6}} & \multicolumn{1}{r|}{\textbf{25.6}} \\
    \multicolumn{1}{|c||}{25} & \multicolumn{1}{r}{61.0} & \multicolumn{1}{r||}{\textbf{99.4}}  & \multicolumn{1}{r}{54.1}          & \multicolumn{1}{r||}{\textbf{61.0}}  & \multicolumn{1}{r}{45.8} & \multicolumn{1}{r||}{\textbf{47.0}}  & \multicolumn{1}{r}{29.7} & \multicolumn{1}{r||}{\textbf{91.0}}  & \multicolumn{1}{r}{50.3}           & \multicolumn{1}{r||}{\textbf{89.9}}            & \multicolumn{1}{r}{ 6.9} & \multicolumn{1}{r||}{\textbf{34.7}}  & \multicolumn{1}{r}{43.8   }       & \multicolumn{1}{r|}{\textbf{51.7}} \\
    \multicolumn{1}{|c||}{30} & \multicolumn{1}{r}{70.7} & \multicolumn{1}{r||}{\textbf{99.4}}  & \multicolumn{1}{r}{65.4}          & \multicolumn{1}{r||}{\textbf{79.4}}  & \multicolumn{1}{r}{58.5} & \multicolumn{1}{r||}{\textbf{59.4}}  & \multicolumn{1}{r}{45.0} & \multicolumn{1}{r||}{\textbf{96.5}}  & \multicolumn{1}{r}{81.0}           & \multicolumn{1}{r||}{\textbf{98.7}}            & \multicolumn{1}{r}{10.9} & \multicolumn{1}{r||}{\textbf{46.4}}  & \multicolumn{1}{r}{50.2   }       & \multicolumn{1}{r|}{\textbf{58.7}} \\
    \multicolumn{1}{|c||}{45} & \multicolumn{1}{r}{93.4} & \multicolumn{1}{r||}{\textbf{99.7}}  & \multicolumn{1}{r}{90.1}          & \multicolumn{1}{r||}{\textbf{99.1}}  & \multicolumn{1}{r}{82.0} & \multicolumn{1}{r||}{\textbf{90.4}}  & \multicolumn{1}{r}{86.6} & \multicolumn{1}{r||}{\textbf{98.9}}  & \multicolumn{1}{r}{\textbf{100.0}} & \multicolumn{1}{r||}{\textbf{100.0}}           & \multicolumn{1}{r}{40.2} & \multicolumn{1}{r||}{\textbf{72.1}}  & \multicolumn{1}{r}{\textbf{83.3}} & \multicolumn{1}{r|}{82.3} \\



    \hline
  \end{tabular}
}
\caption{Percentage of total frames in a sequence that have an error of less $x$~mm. We observe that \texttt{SAG} outperforms \texttt{SoG} in all sequences and error bounds.
  The values in bold face indicate the best values for a given error bound.}
\label{tab:error_rate}
\end{table*}

\parahead{Effect of Number of Cameras}
To evaluate the scalability of our method to the number of cameras we  conducted an experiment where each camera was progressively disabled with total active cameras ranging from $2$ to $5$. This leads to $26$ possible camera combinations for each sequence and a total of $156$ runs for both the \texttt{SAG} and \texttt{SoG} methods.
We excluded the \texttt{random} sequence as it was too challenging for tracking with 3 or less cameras.

Figure~\ref{fig:results2} shows the average error over all runs for varying cameras.
Clearly, \texttt{SAG} produces lower errors and standard deviations for all camera combinations.
We also observe a diverging trend and hypothesize that as the number of cameras is increased the gap between \texttt{SAG} and \texttt{SoG} will also increase.
This may be important for applications requiring very precise tracking such as motion capture for movies.
%
We associate the improvements in accuracy of \texttt{SAG} with its ability to approximate the users' hand better than \texttt{SoG}.
Figure~\ref{fig:handmodel} (b, d) visualizes the projected model density and reveals a better approximation for \texttt{SAG}.

\begin{figure*}[!htbp]
  \centering
   \subfigure{\includegraphics[width=0.45\textwidth]{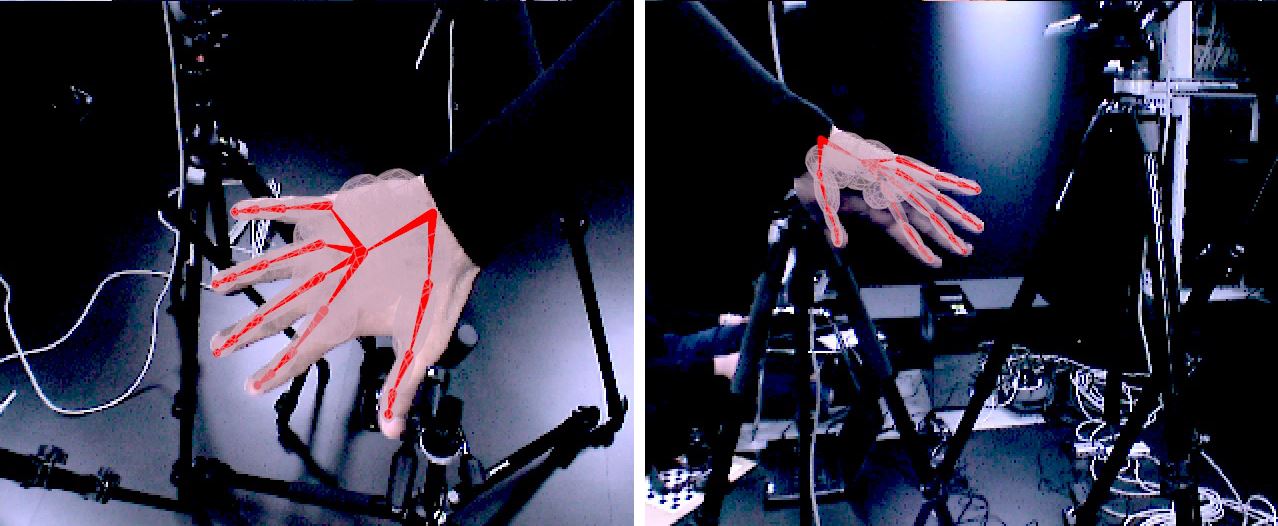}}$\ \ $
   \subfigure{\includegraphics[width=0.45\textwidth]{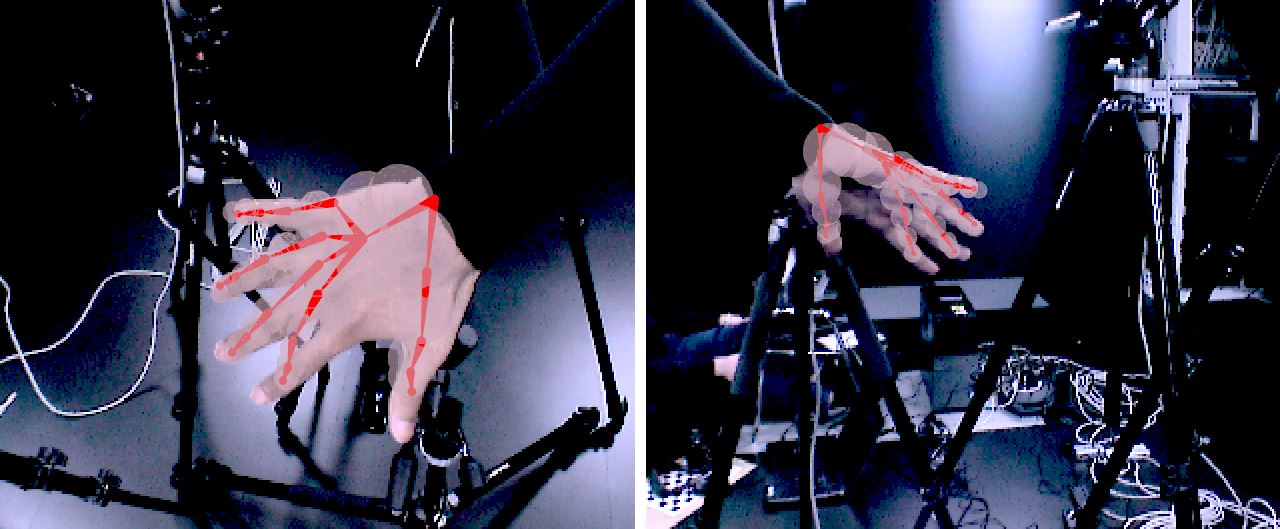}}
   \subfigure{\includegraphics[width=0.45\textwidth]{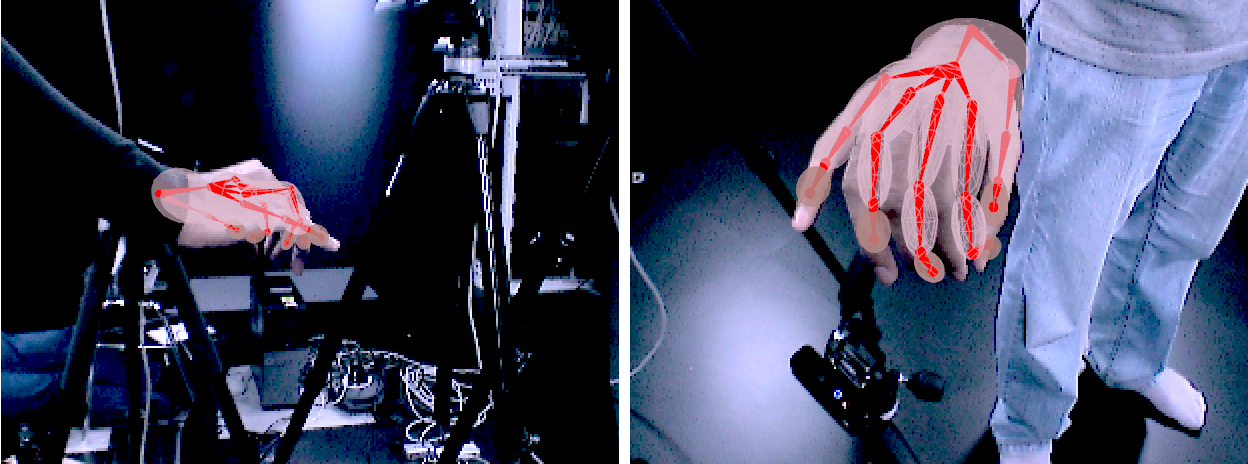}}$\ \ $
   \subfigure{\includegraphics[width=0.45\textwidth]{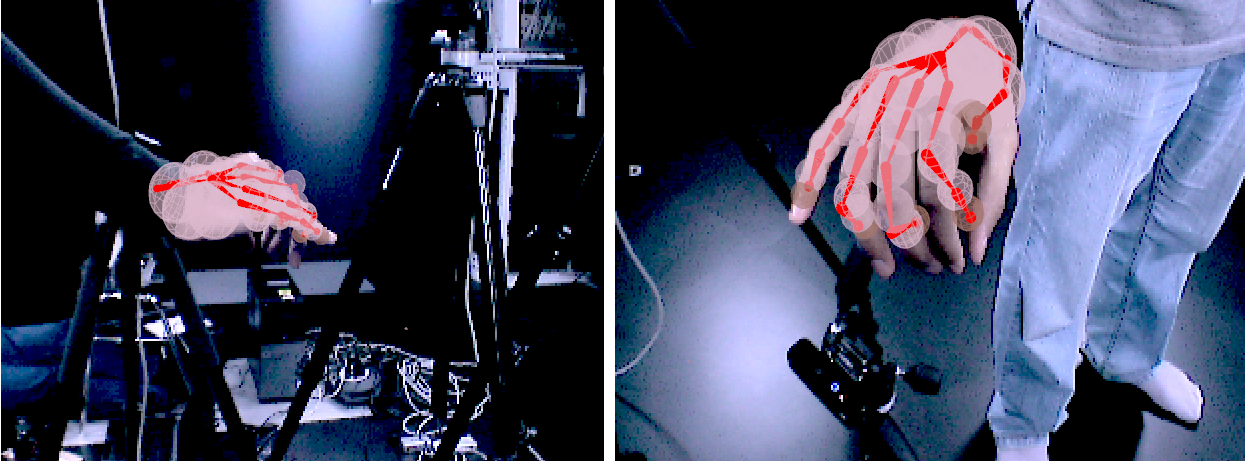}}
   \subfigure{\includegraphics[height=1.25in]{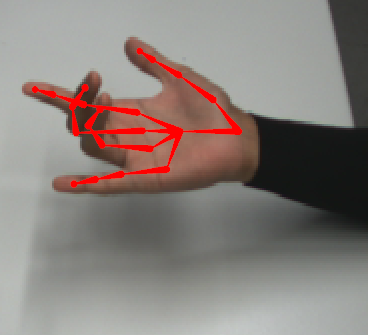}}
   \subfigure{\includegraphics[height=1.25in]{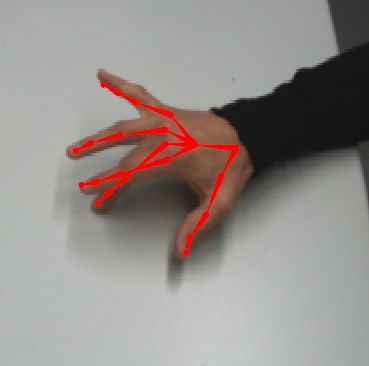}}
   \subfigure{\includegraphics[height=1.25in]{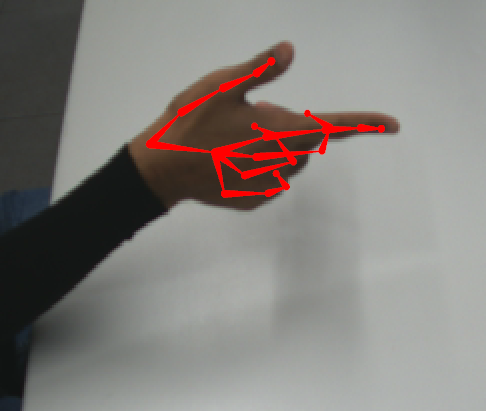}}
   \subfigure{\includegraphics[height=1.25in]{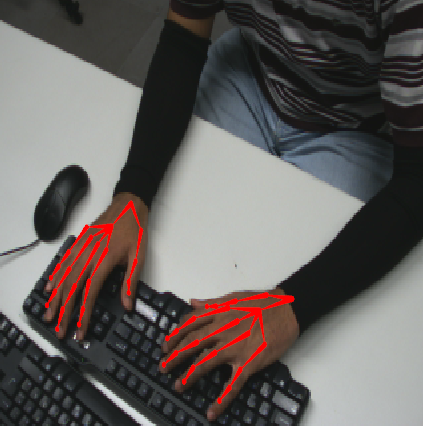}}
   \subfigure{\includegraphics[height=1.25in]{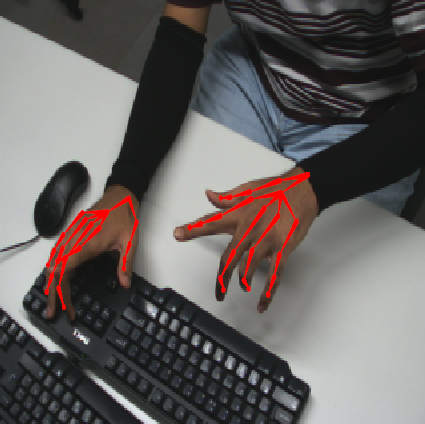}}
   \caption{\textbf{First Two Rows}: Comparison of \texttt{SAG} (left) and \texttt{SoG} (right) for two frames in the Dexter 1 dataset.
     In the first row, \texttt{SAG} covers the hand much better during a fast motion of the hand in spite of using fewer primitives.
     In the second row, a challenging motion is performed for which \texttt{SAG} performs better.
     \textbf{Bottom Row}: Realtime tracking results for one hand with different actors, and two hands. Please see supplementary video for results from hand + object tracking.}
   \label{fig:qual}
\end{figure*}

\parahead{Qualitative Tracking Results}
Finally, we show several qualitative results of tracking in Figure~\ref{fig:qual} comparing \texttt{SAG} and \texttt{SoG}.
Since our tracking approach is flexible we are also able to track additional simple objects such as a plate using only a few primitives.
These additional results, qualitative comparison to \cite{tang_latent_2014}, and failure cases can be found in the supplementary video.

\section{Discussion and Future Work}
As demonstrated in the above experiments our method advances state of the art methods in accuracy
and is suitable for real-time applications.
However, the generative method can lose tracking because of fast hand motions.
Like other hybrid methods, we could augment our method with a discriminative tracking strategy.
The generality of our method allows easy integration into such a hybrid framework.

In terms of utility, we require the user to wear a black sleeve and we use multiple calibrated cameras.
These limitations could be overcome if we would only rely the depth data for our tracking.
Since the SAG representation is data agnostic, we could model the depth image as a SAG as well.
We intend to explore these improvements in the future.

\section{Conclusion}
We presented a method for articulated hand tracking that uses a novel
Sum of Anisotropic Gaussians (SAG) representation to track hand motion.
Our SAG formulation uses a full perspective projection model and uses only a few Gaussians to model the hand.
Because of our smooth and differentiable pose fitting energy, we are able to perform fast gradient-based pose
optimization to achieve real-time frame rates.
Our approach produces more robust and accurate tracking than previous methods
while featuring advantageous numerical properties and comparable runtime.
We demonstrated our accuracy and robustness on standard datasets by comparing with relevant work from literature.

\small{
\parahead{Acknowledgments}
This work was supported by the ERC Starting Grant CapReal.
We would like to thank James Tompkin and Danhang Tang.
}

{\small
\bibliographystyle{ieee}
\bibliography{content/eccv2014}

\begin{thebibliography}{10}\itemsep=-1pt

\bibitem{athitsos_estimating_2003}
V.~Athitsos and S.~Sclaroff.
\newblock Estimating 3d hand pose from a cluttered image.
\newblock In {\em Proc.\ of {IEEE} CVPR 2003}, pages 432--9 vol.2.

\bibitem{baak_data-driven_2011}
A.~Baak, M.~Muller, G.~Bharaj, H.-P. Seidel, and C.~Theobalt.
\newblock A data-driven approach for real-time full body pose reconstruction
  from a depth camera.
\newblock In {\em Proc.\ of {ICCV} 2011}, pages 1092 --1099.

\bibitem{ballan_motion_2012}
L.~Ballan, A.~Taneja, J.~Gall, L.~Van~Gool, and M.~Pollefeys.
\newblock Motion capture of hands in action using discriminative salient
  points.
\newblock In {\em Proc.\ of {ECCV} 2012}, pages 640--653 v.7577.

\bibitem{eberly_perspective_1999}
D.~Eberly.
\newblock Perspective projection of an ellipsoid.
\newblock \url{http://www.geometrictools.com/}, 1999.

\bibitem{erol_vision-based_2007}
A.~Erol, G.~Bebis, M.~Nicolescu, R.~D. Boyle, and X.~Twombly.
\newblock Vision-based hand pose estimation: A review.
\newblock {\em Computer Vision and Image Understanding}, pages 52--73 v.108,
  2007.

\bibitem{hamer_object-dependent_2010}
H.~Hamer, J.~Gall, T.~Weise, and L.~Van~Gool.
\newblock An object-dependent hand pose prior from sparse training data.
\newblock In {\em Proc.\ of {CVPR} 2010}, pages 671--678.

\bibitem{hamer_tracking_2009}
H.~Hamer, K.~Schindler, E.~Koller-Meier, and L.~Van~Gool.
\newblock Tracking a hand manipulating an object.
\newblock In {\em Proc.\ of {ICCV} 2009}, pages 1475--1482.

\bibitem{keskin_hand_2012}
C.~Keskin, F.~Kıraç, Y.~E. Kara, and L.~Akarun.
\newblock Hand pose estimation and hand shape classification using
  multi-layered randomized decision forests.
\newblock In {\em Proc.\ of {ECCV} 2012}, pages 852--863.

\bibitem{krivic_contour_2003}
J.~Krivic and F.~Solina.
\newblock Contour based superquadric tracking.
\newblock Lecture Notes in Computer Science, pages 1180--1186. 2003.

\bibitem{lin_modeling_2000}
J.~Lin, Y.~Wu, and T.~Huang.
\newblock Modeling the constraints of human hand motion.
\newblock In {\em Workshop on Human Motion, 2000. Proceedings}, pages 121
  --126.

\bibitem{maccormick_partitioned_2000}
J.~MacCormick and M.~Isard.
\newblock Partitioned sampling, articulated objects, and interface-quality hand
  tracking.
\newblock In {\em Proc.\ of {ECCV} 2000}, pages 3--19.

\bibitem{oikonomidis_efficient_2011}
I.~Oikonomidis, N.~Kyriazis, and A.~Argyros.
\newblock Efficient model-based 3d tracking of hand articulations using kinect.
\newblock In {\em Proc.\ of BMVC 2011}, pages 101.1--101.11.

\bibitem{oikonomidis_full_2011}
I.~Oikonomidis, N.~Kyriazis, and A.~Argyros.
\newblock Full {DOF} tracking of a hand interacting with an object by modeling
  occlusions and physical constraints.
\newblock In {\em Proc.\ of {ICCV} 2011}, pages 2088--2095.

\bibitem{qian_realtime_2014}
C.~Qian, X.~Sun, Y.~Wei, X.~Tang, and J.~Sun.
\newblock Realtime and robust hand tracking from depth.
\newblock In {\em Proc.\ of {CVPR} 2014}.

\bibitem{rehg_visual_1994}
J.~Rehg and T.~Kanade.
\newblock Visual tracking of high {DOF} articulated structures: An application
  to human hand tracking.
\newblock In {\em Proc.\ of {ECCV} 1994}, volume 801, pages 35--46 v.801.

\bibitem{romero_hands_2010}
J.~Romero, H.~Kjellstrom, and D.~Kragic.
\newblock Hands in action: real-time 3d reconstruction of hands in interaction
  with objects.
\newblock In {\em Proc.\ of {IEEE ICRA} 2010}, pages 458--463.

\bibitem{rosales_combining_2006}
R.~Rosales and S.~Sclaroff.
\newblock Combining generative and discriminative models in a framework for
  articulated pose estimation.
\newblock {\em International Journal of Computer Vision}, pages 251--276 v.67,
  2006.

\bibitem{shotton_real-time_2011}
J.~Shotton, A.~Fitzgibbon, M.~Cook, T.~Sharp, M.~Finocchio, R.~Moore,
  A.~Kipman, and A.~Blake.
\newblock Real-time human pose recognition in parts from single depth images.
\newblock In {\em Proc.\ of {CVPR} 2011}, pages 1297--1304.

\bibitem{sridhar_interactive_2013}
S.~Sridhar, A.~Oulasvirta, and C.~Theobalt.
\newblock Interactive markerless articulated hand motion tracking using {RGB}
  and depth data.
\newblock In {\em Proc.\ of {ICCV} 2013}.

\bibitem{stenger_model-based_2006}
B.~Stenger, A.~Thayananthan, P.~H.~S. Torr, and R.~Cipolla.
\newblock Model-based hand tracking using a hierarchical bayesian filter.
\newblock {\em {IEEE TPAMI}}, 28:1372--1384, 2006.

\bibitem{stoll_fast_2011}
C.~Stoll, N.~Hasler, J.~Gall, H.~Seidel, and C.~Theobalt.
\newblock Fast articulated motion tracking using a sums of gaussians body
  model.
\newblock In {\em Proc.\ of {ICCV} 2011}, pages 951 --958.

\bibitem{tang_latent_2014}
D.~Tang, H.~J. Chang, A.~Tejani, and T.-K. Kim.
\newblock Latent regression forest: Structured estimation of 3d articulated
  hand posture.
\newblock In {\em Proc.\ of {CVPR} 2014}.

\bibitem{tang_real-time_2013}
D.~Tang, T.-H. Yu, and T.-K. Kim.
\newblock Real-time articulated hand pose estimation using semi-supervised
  transductive regression forests.
\newblock In {\em Proc.\ of {ICCV} 2013}.

\bibitem{wang_6d_2011}
R.~Wang, S.~Paris, and J.~Popovi\'c.
\newblock 6d hands: markerless hand-tracking for computer aided design.
\newblock In {\em Proc.\ of {ACM UIST} 2011}, pages 549--558.

\bibitem{wang_real-time_2009}
R.~Y. Wang and J.~Popovi\'c.
\newblock Real-time hand-tracking with a color glove.
\newblock {\em {ACM TOG}}, page 63:1–63:8 v.28, 2009.

\bibitem{wei_accurate_2012}
X.~Wei, P.~Zhang, and J.~Chai.
\newblock Accurate realtime full-body motion capture using a single depth
  camera.
\newblock {\em {ACM TOG}}, pages 188:1--188:12 v.31, 2012.

\bibitem{wendland_piecewise_1995}
H.~Wendland.
\newblock Piecewise polynomial, positive definite and compactly supported
  radial functions of minimal degree.
\newblock {\em Adv Comput Math}, 4(1):389--396, 1995.

\bibitem{wu_capturing_2001}
Y.~Wu, J.~Lin, and T.~Huang.
\newblock Capturing natural hand articulation.
\newblock In {\em Proc.\ of {ICCV} 2001}, volume~2, pages 426--432 vol.2.

\bibitem{xu_efficient_2013}
C.~Xu and L.~Cheng.
\newblock Efficient hand pose estimation from a single depth image.
\newblock In {\em Proc.\ of {ICCV} 2013}.

\bibitem{ye_accurate_2011}
M.~Ye, X.~Wang, R.~Yang, L.~Ren, and M.~Pollefeys.
\newblock Accurate 3d pose estimation from a single depth image.
\newblock In {\em Proc.\ of {ICCV} 2011}, pages 731--738.

\end{thebibliography}
}

\end{document}